\documentclass[fleqn,11pt]{llncs}

\usepackage{url}
\usepackage{tikz}
\usepackage{amsmath}
\usepackage{amsfonts}
\usepackage{listings}
\usepackage{numprint}
\usepackage{tabularx}
\usepackage{longtable}
\usepackage{subcaption}
\usepackage[utf8]{inputenc}

\newcolumntype{R}[1]{>{\raggedleft\let\newline\\\arraybackslash\hspace{0pt}}m{#1}}
\newcolumntype{L}[1]{>{\raggedright\let\newline\\\arraybackslash\hspace{7pt}}m{#1}}

\usetikzlibrary{shapes,arrows,matrix}

\tikzstyle{block} = [rectangle, draw, fill=green!20,
text width=12em, text centered, rounded corners, minimum height=3.5em]
\tikzstyle{sblock} = [rectangle, draw, fill=green!20,
text width=8em, text centered, rounded corners, minimum height=2em]
\tikzstyle{final} = [rectangle, draw, fill=green!20,
text width=25em, text centered, rounded corners, minimum height=3.5em]
\tikzstyle{stage} = [rectangle, draw, fill=red!20,
text centered, rounded corners, minimum height=3em, text width=5.5cm]
\tikzstyle{title} = [rectangle, draw, fill=blue!20,
text centered, rounded corners, minimum height=3em, text width=14cm]
\tikzstyle{line} = [draw, -latex']

\newcommand{\gname}[1]{\mathsf{#1}}
\newcommand{\gT}{\gname{T}}

\newcommand{\gG}{\gname{G}}
\newcommand{\gH}{\gname{H}}

\newcounter{nalg}[section]
\renewcommand{\thenalg}{\thesection .\arabic{nalg}}
\DeclareCaptionLabelFormat{algocaption}{Algorithm \thenalg}

\lstnewenvironment{algorithm}[1][]
{   
	\refstepcounter{nalg} 
	\captionsetup{labelformat=algocaption,labelsep=colon} 
	\lstset{ 
		mathescape=true,
		frame=tB,
		numbers=left, 
		numberstyle=\tiny,
		breaklines=true,
		postbreak=\mbox{\textcolor{red}{$\hookrightarrow$}\space},
		basicstyle=\ttfamily\small, 
		keywordstyle=\color{black}\bfseries,
		keywords={,input, output, return, datatype, function, if, else, foreach, while, begin, end, }
		numbers=left,
		xleftmargin=.04\textwidth,
		tabsize=2,
		#1 
	}
}
{}

\title{A Top-down Supervised Learning Approach to Hierarchical Multi-label
	Classification in Networks}

\author{
	Miguel Romero \and
	Jorge Finke \and
	Camilo Rocha
}

\institute{
	Department of Electronics and Computer Science \\
	Pontificia Universidad Javeriana, Cali, Colombia
	\email{\{miguelangel.romero,jfinke,camilo.rocha\}@javerianacali.edu.co}
}

\begin{document}

\maketitle

\begin{abstract}
	Node classification is the task of inferring or predicting missing
  node attributes from information available for other nodes in a
  network. This paper presents a general prediction model to
  \emph{hierarchical} multi-label classification (HMC), where the
  attributes to be inferred can be specified as a strict poset.
	It is based on a top-down classification approach that addresses
	hierarchical multi-label classification with supervised learning by
	building a local classifier per class.
 	The proposed model is showcased with a case study on the prediction
  of gene functions for \textit{Oryza sativa Japonica}, a variety of
  rice.
	It is compared to the Hierarchical Binomial-Neighborhood, a
	probabilistic model,
	by evaluating both approaches in terms of prediction performance and
  computational cost. The results in this work support the working
  hypothesis that the proposed model can achieve good levels of
  prediction efficiency, while scaling up in relation to the state of
  the art.
\end{abstract}

\section{Introduction}
\label{sec:intro}

Network representations provide a formal framework to specify relationships
between interconnected entities (nodes). In a number of scenarios, such
frameworks annotate nodes with attributes to naturally identify groups whose
members are related by, e.g., a particular similarity measure.  In analyzing
networks with node attributes, most studies assume that a node can take a
finite number of possible values, each of which is called a \emph{class}. A
class may represent the gender of a user in a social network or the function
associated to a protein in a protein-protein interaction network.

\subsection*{The Problem}

The task of inferring or predicting missing node attributes from
information available for other nodes in a network is called
\emph{node classification} (also referred to as \emph{attribute
	prediction})~\cite{bhagat-nodecl-2011}. Several formulations to the
node classification problem have been proposed over the past
decades. The problem of inferring an attribute from exactly one of two
classes is referred to as \emph{binary}
classification~\cite{khan-binary-2010}. The extension of the binary
classification problem to \emph{any} finite (non-zero) number of
classes is referred to as \emph{multi-class}
classification~\cite{mills-multiclass-2021}. Furthermore, each type
of problem is categorized as a \emph{multi-label} classification, if a
node is allowed to be simultaneously associated to more than one
class~\cite{prajapati-multilabel-2012}.

For most techniques that address the above-mentioned problems, node
classification is generally carried out independently for each
class~\cite{abu-ngcn-2019,bhagat-nodecl-2011,hamilton-induct-2017,kipf-gcn-2017,xiao-gnn-2021}.
The main limitation of such compartmentalized approaches is
that they ignore hidden relationships among classes, even when certain
class relationships may serve as an input to improve the accuracy of
attribute prediction.

In practice, classes may have explicit relations specifying their
dependencies. For example, this is the case with the Gene Ontology hierarchy
because genes and proteins associated to a function must also be associated to
the ancestors of such function.
The authors in~\cite{silla-hierarchy-2011}, amid this limitation,
define dependencies between classes as ancestral relations by means of
a hierarchy represented by a directed acyclic graph.
A connection from class $C_1$ to class $C_2$ in the hierarchy means
that every node with attribute $C_1$ also has attribute $C_2$ (i.e.,
the nodes having attribute $C_1$ are a subclass of the nodes having
attribute $C_2$).

Formally, a classification problem is considered \emph{hierarchical}
if and only if its hierarchy of classes is a strict partial order
(i.e., a \emph{strict poset} or, equivalently, a directed acyclic
graph).
A strict poset $(C,\prec)$ over a finite set $C$ of classes defines a
binary relation $\prec$ on $C$ that is asymmetric, anti-reflexive, and
transitive.
For instance, the hierarchy of biological processes can be defined
over a strict poset according to which the functions \emph{cell
	death}, \emph{programmed cell death}, and \emph{apoptotic process}
are ordered by $\textit{cell death} \prec \textit{programmed cell
	death} \prec \textit{apoptotic process}$~\cite{go-go-2019}.
Note that the transitive property of the order $\prec$ guarantees that
\emph{cell death} is also the ancestor of \emph{apoptotic
	process}. Note also that since the strict poset is anti-reflexive, no
process can be ancestor of itself. Finally, the asymmetry property
guarantees that \emph{apoptotic process} cannot be ancestor of
\emph{programmed cell death}.
Indeed, any strict poset is isomorphic to a directed acyclic graph
(DAG), a concept more closely related to graph and network
  analysis.

Given a graph with some labeled nodes (i.e., nodes associated to classes) and
a class hierarchy, the expected outcome of a hierarchical classification
problem is a collection of predicted associations between nodes and classes.
An \emph{inconsistent prediction} for a hierarchical multi-label
classification problem refers to the fact that a node is inferred to
have a particular class $C$, but the outcome of the classifier fails
to infer the node's association to all ancestor classes of $C$.  In
other words, an inconsistent prediction states that the prediction
does not satisfy the ancestral relations for some class $C$. In many
scenarios, it is desirable to rule out inconsistent prediction: that
is, if a classifier predicts a particular class $C$ for a node, then
it should also predict all the ancestors of $C$ for that node;
conversely, if a classifier does not predict $C$ for a node, then it
should not predict any of $C$'s descendants for that node.  This
constraint is often referred to as the \textit{true-path rule} in Gene
Ontology~\cite{ashburner-go-2000,valentini-tpr-2009}.

The efforts to classify nodes generally aim to comply with the
underlying ancestral relations between classes in scenarios where such
a hierarchy is known (and thus avoid inconsistent
prediction). Consider a social network where nodes represent
individuals and node attributes represent different levels of
education. If a predictor outputs an individual without an
undergraduate degree as candidate to have graduate degree, it fails to
comply with the hierarchical organization of educational levels, in
many reasonable scenarios, thereby violating the true-path rule.

Hierarchical classification problems may further be categorized
depending on the approach used for training the underlying model. On
the one hand, a \textit{top-down} approach involves a binary
classifier for each class in the hierarchy. In this case, the
classifier associated to each class is trained iteratively from the
roots (i.e., the classes without ancestors in the hierarchy) to the
leaves (i.e., the classes without descendants in the hierarchy). In
addition, local information about the ancestors and descendants of a
class in the hierarchy is used to avoid independent predictions. On
the other hand, a \textit{big-bang} approach involves a multi-label
classifier that considers the entire hierarchy of ancestral relations
at once. The multi-label classifier is trained just once with the
information of every class in the hierarchy and its dependencies.

\subsection*{Related Work}

Several studies have applied both top-down and big-bang approaches
across different
domains~\cite{jiang-hbn-2008,dimitrovski-hmlc-2010,bi-hmlc-2011,ramirez-corona-hmlc-2016}.
The authors in~\cite{jiang-hbn-2008} proposes a top-down approach,
called Hierarchical Binomial-Neighborhood (HBN), to predict protein
functions in yeast \textit{Saccharomyces cerevisiae}. It is shown by
the authors that the hierarchical structure of functions can be
exploited to completely avoid inconsistent predictions and, at the
same time, outperform approaches based on independent class
prediction. However, they point out that the main limitation of their
approach is the high computational effort required for assigning
probability weights to every protein-function pair.
The authors in~\cite{ramirez-corona-hmlc-2016} introduce a top-down
approach based on Chained Path Evaluation (CPE), which uses a
classifier to train each non-leaf class (i.e., each class with at
least one descendant) in the hierarchy. Information on ancestral
relations is included in the classifier by adding an extra feature
with the prediction of parents of each class. As
in~\cite{jiang-hbn-2008}, the computational cost of the CPE model
grows exponentially as a function of the number of paths in the
hierarchy.  The use of big-bang approaches is, in general, also
limited by their high computational
demands. In~\cite{dimitrovski-hmlc-2010}, for example, the authors
present a big-bang approach that addresses hierarchical multi-label
classification based on Predictive Clustering Trees (PCTs). The
computational cost of the PCT approach is directly proportional to the
size of the hierarchy.

Other studies address the node classification problem and obtain
state-of-the-art performance for different case studies (see,
e.g.,~\cite{abu-ngcn-2019,chen-gfp-2021,hamilton-induct-2017,kipf-gcn-2017,makrodimitris-afp-2020,xiao-gnn-2021}). However, they do not take into account dependencies between classes (hierarchical or not), for they focus on multi-class instead of multi-label problems. For this reason, such developments can not be compared directly to assess hierarchical multi-label classification prediction.

\subsection*{Main Contribution}

This work introduces a top-down classification approach that addresses
hierarchical multi-label classification (HMC) using supervised learning. 
Given a network $\gG =(V_{\gG}, E_{\gG})$, an assignment of classes to nodes
in the network, and a class hierarchy specified as a directed acyclic graph
$\gH =(V_{\gH}, E_{\gH})$, the hierarchical multi-label classification problem
is addressed by building a binary classifier for each class. Classifiers are
built iteratively from the roots of the hierarchy to the leaves.
The approach uses a correction mechanism to guarantee that the
true-path rule is satisfied by the classifier's outcome; it is
enforced by computing cumulative probabilities along the paths of
classes in the input hierarchy.

The results in this work support the working hypothesis that the
proposed approach can achieve good levels of prediction efficiency,
while scaling up in relation to the state of the art. This approach is
showcased with a case study on the prediction of gene functions for
\textit{Oryza sativa Japonica}, a variety of rice. It is compared to
the probabilistic HBN model~\cite{jiang-hbn-2008}, by evaluating
both approaches in terms of prediction performance (by means of the
true positive and true negative rates) and in terms of their
computational cost (by means of a comparison of the execution
time). In the case study, the prediction task uses two inputs. Namely,
(i) a gene co-expression network (GCN), in which a node represents a
gene and a class of a node represents a gene function; and (ii) the
hierarchical structure of biological processes defined
in~\cite{go-go-2019}. The goal of the prediction task is to infer gene
attributes from 15 sub-hierarchies grouping \numprint{1938} biological
processes associated to \numprint{19663} genes.

\textbf{Outline.} The remainder of the paper is organized as follows.
Section~\ref{sec:model} introduces the approach for node classification
where classes have a hierarchical organization. Section~\ref{sec:gene}
describes the problem of predicting gene functions. It also presents the
results of applying the proposed model to \textit{Oryza sativa Japonica}.
Finally, Section~\ref{sec:concl} draws some concluding remarks and future
research directions.

\section{Hierarchical Classification}
\label{sec:model}

This section presents a top-down classification approach in the form
of a supervised learning model for hierarchical multi-label
classification.

The input of the model are
a graph $\gG =(V_{\gG}, E_{\gG})$ specifying an
undirected network with nodes $V_{\gG}$ and edges $E_{\gG}$,
a directed
acyclic graph $\gH =(V_{\gH}, E_{\gH})$, with vertices $V_{\gH}$ and edges $E_{\gH}$
disjoint from $V_{\gG}$ and $E_{\gG}$, respectively, representing the
hierarchy of classes, and
a function $\phi : V_G \mapsto 2^{V_{\gH}}$ with a partial assignment of
classes to nodes in the network. 
For $v \in V_{\gG}$, the set $\phi(v) \subseteq V_{\gH}$ is the collection of
classes initially associated to $v$.
It is assumed that $\phi$ satisfies the true-path rule for the hierarchy
$\gH$, meaning that if a node $v$ satisfies $C \in \phi(v)$ for a class $C \in
V_{\gH}$, then $\phi(v)$ must contain all the ancestors of $C$ in $\gH$.
As mentioned in the introduction, the DAG $\gH$ uniquely represents a
strict poset.  The goal of the model is then to build a function
$\phi' : V_G \mapsto 2^{V_{\gH}}$ extending $\phi$ with new
assignments of nodes in $V_{\gG}$ to classes in $V_{\gH}$.
Figure~\ref{fig:input}A depicts an example of the input of the model where the
nodes of the network $\gG$ are labeled with classes $A$-$E$ and the hierarchy
of classes $\gH$ is a DAG. According to the true-path rule, nodes labeled with
class $E$ are also related to classes $A$, $B$, and $C$. The objective is to
predict new associations between nodes and classes for either nodes with or
without labels.

The rest of this section is devoted to describe the main steps behind
the construction of the supervised learning model.

\subsection{Hierarchy Normalization}

Hierarchies are represented as directed acyclic graphs where, in general,
nodes can have any (finite) number of parents.  Since the approach presented
in this work assumes that every node has at most one parent, a topological
traversal algorithm for directed graphs (see,
e.g.,~\cite{knuth-programming-1997}) is used to transform $\gH$ into a tree,
when required. In this way, the resulting model can take as input any
hierarchy.

This algorithm uses the structure of $\gH$ (not its tree version) and
its distribution of classes. Given an ancestral relation $A \to B$
(i.e., class $A$ is a direct ancestor of class $B$), a weight $w(A,B)$
for such an edge is defined as the ratio between the number of nodes
associated to $B$ (i.e., the size of the set $\phi^{-1}(B)$) and the
number of nodes associated to $A$ (i.e., the size of the set
$\phi^{-1}(A)$).
Since all nodes associated to $B$ must be associated to $A$, then by
definition each weight $w(A,B)$ is in the range $[0,1]$.
For any node $B$ with $n\geq 1$ parents $A_1,\ldots,A_n$ in $\gH$
(i.e., $A_i \to B$ in $E_{\gH}$, for $1 \leq i \leq n$), the parent of $B$
in the resulting tree is the node $A_j$ maximizing $w(A_j, B)$ among
all the $A_i$'s. Ties are broken arbitrarily. This process can be
effectively computed in time and space $O(|V_{\gH}| + |E_{\gH}|)$, namely, in
resources linear in the size of $\gH$. Such a process, based on a
topological-sorting traversal, is described in
Algorithm~\ref{alg:top}. 
Finally, note that the topological sorting
of the vertices of $\gH$ in Algorithm~\ref{alg:top}, can be exploited
to compute the value of function $w(\_,\_)$ by dynamic programming in
space $\Theta(|V_{\gH}|)$.  More precisely, a function $\rho : V_{\gH} \to
\mathbb{N}$ assigning to each class $B_i$ its number of descendants
$\rho(B_i)$ in $\gH$ can be computed from the direct descendants of
$B_i$, which are processed before $B_i$ in the topological sorting of
$V_{\gH}$.

\begin{algorithm}[caption={Topological-sorting based traversal for hierarchy normalization}, label={alg:top}]
	input: $\gH$
	output: $\gT = (V_{\gH}, E_{\gT})$
	set $E_{\gT} = \{\}$
	compute a topological sorting $B_1, \ldots, B_n$ of $V_{\gH}$ (leaves first)
	foreach class $B_i$ with $1\leq i\leq n$
	  identify all parents $A_1, \dots, A_m$ of $B_i$ in $\gH$
	  foreach parent $A_j$ of $B_i$ with $1\leq j\leq m$
	    compute the weight $w(A_j, B_i)$
 	  identify $A_k$ with $w(A_k, B_i)\leq w(A_j, B_i)$ for $1\leq j\leq m$
	  extend $E_{\gT}$ with $(A_k \to B_i)$ 
	return the tree $\gT$
\end{algorithm}

\begin{figure}[tbph!]
	\centering
	\includegraphics[width=0.8\linewidth]{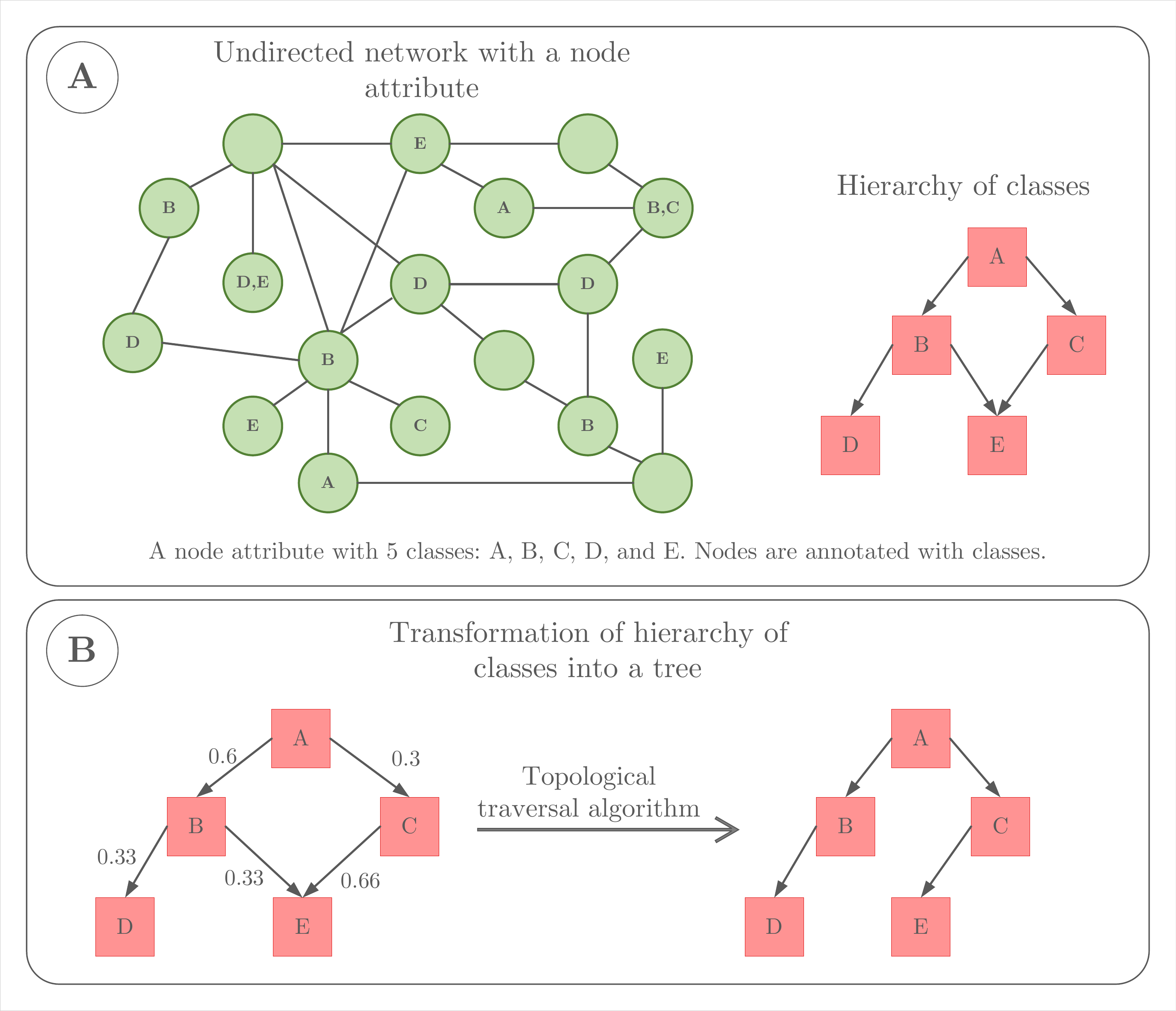}
	\caption{A. The classification approach gets as input a network with a node
		attribute and a set of known association between nodes and classes, and the
		hierarchy of ancestral relations represented as a DAG. Note that there are
		more nodes associated to class $B$ than $C$. B. The DAG representation of
		the hierarchy is transformed into a tree using a topological traversal
		algorithm, based on the distribution of the classes in the network.
		Since classes $B$ and $C$ are ancestors of $E$, the ratio of
		nodes associated to $E$ and $C$ is higher that the ratio for
		$E$, and $B$ ($w(C,E)>w(B,E)$), the algorithm removes edge
		$(B,E)$ and returns a tree.}
	\label{fig:input}
\end{figure}

As an example, consider the hierarchy depicted in Figure~\ref{fig:input}B.
Note that class $E$ has more than one parent, there exists an ancestral
	relation from $B$ to $E$ and from $C$ to $E$ (i.e, $B \to E$ and $C \to E$,
respectively). By the true-path rule, nodes associated to class $D$ are also
associated to class $B$, and the ones associated to class $E$ are associated
to both $B$ and $C$. Since there are 4 nodes associated to $E$, 4 to $D$, 4 to
$B$, and $2$ to $C$, the weight $w(B, E)$ is $0.33$ and the weight $w(C, E)$
is $0.66$. Therefore, the topological-sorting traversal will remove edge $B
\to E$.

In the rest of this paper, it is assumed that hierarchy $\gH$ is
indeed a tree $\gT$.

\subsection{The Model}

\begin{figure}[tbph!]
	\centering
	\resizebox{.95\textwidth}{!}{
		\begin{tikzpicture}[node distance = 1.5cm, auto, ampersand replacement=\&]
		\node [title] (wflow) {
			\LARGE{Hierarchical classification model} \break
			\normalsize{\textbf{Inputs:}}
			\footnotesize{an undirected network, and a
				DAG that represents the hierarchy of classes (including the assignments of classes to some nodes)}};
		\node [stage, below of=wflow, node distance=2.1cm] (prepr) {
			\underline{\small Data pre-processing}
			\break \footnotesize{Processing of input data for dataset building}};
		\matrix[draw, column sep=0.5cm, row sep=0.5cm, below of=prepr, fill=gray!20, rounded corners, node distance=2.3cm, inner sep=0.2cm] (premat) {
			\node [block] (chier) {
				\underline{\small Hierarchies retrieving} \break
				\footnotesize{Compute hierarchies from ancestral relations of classes}}; \&
			\node [block] (cprop) {
				\underline{\small Dataset building} \break
				\footnotesize{Compute topological properties of the network}}; \\
		};
		\node [stage, below of=premat, node distance=2.3cm] (mlclf) {
			\underline{\small Hierarchical classification} \break
			\footnotesize{Top-down strategy, built a binary classifier per class}};
		\matrix[draw, column sep=0.5cm, row sep=0.5cm, below of=mlclf, fill=gray!20, rounded corners, node distance=3.6cm, inner sep=0.2cm] (clfmat) {
			\node [block] (kfold) {
				\underline{\small \textit{k}-fold}\break \underline{\small cross-validation} \break\footnotesize{Split dataset into \textit{k} folds}}; \& \node [block] (smote) {
				\underline{\small SMOTE} \break \footnotesize{Over-sampling of minority class}};\\
			\node [block] (xgb) {
				\underline{\small Classification} \break \footnotesize{Training the binary classifier}}; \& \node [block] (hpt) {
				\underline{\small Hyper-parameter}\break \underline{\small tuning} \break \footnotesize{Optimize hyper-parameter of the classifier}};\\
		};
		\node [sblock, below of=clfmat, node distance=3.2cm] (pred) {\small Prediction};
		\node [stage, below of=pred, node distance=2cm] (eval) {
			\underline{\small Performance evaluation} \break
			\footnotesize{Computing recall, precision, and F1 scores, and confusion matrix}};
		\path [line,dashed] (wflow) -- (prepr);
		
		\path [line] (prepr) -- (premat);
		\path [line] (chier) -- (cprop);
		\path [line] (premat) -- (mlclf);
		\path [line] (kfold) -- (smote);
		\path [line] (smote) -- (hpt);
		\path [line] (hpt) -- (xgb);
		
		\path [line] (mlclf) -- (clfmat);
		\path [line] (clfmat) -- (pred);
		\path [line] (pred) -- (eval);
		
		\path [line, thick] (pred.east) -- ++(4.3,0) -- ++(0,6.8) -- node[yshift=0.7cm, xshift=0.3cm, text width=8em, font=\footnotesize] {*prediction from ancestor} (mlclf.east);
		
		\path [line] (pred.south) -- ++(0,-0.37) -- ++(-2.1,0) -- ++(0,0.73) node[xshift=-1.2cm,yshift=-.35cm,font=\footnotesize\bfseries, text width=7em] {Cumulative probabilities} -- (pred.west);		
		\end{tikzpicture}
	}
	\caption{Framework of the hierarchical multi-label classification
		approach. The approach is split into three stages: data
		pre-processing, class prediction and performance evaluation. The
		approach is applied for every resulting sub-hierarchy $\gH'$
		independently. Ancestral relations between classes are included in
		the model as features with the prediction of ancestors and are
		represented by the upward arrow in the prediction stage.
		In addition, a correction mechanism for inconsistencies is included by
		means of cumulative probabilities, which are computed according to the
		path of classes in the sub-hierarchy. If the probability of association
		between a node and a class is close to zero, then the cumulative
		probability of the association between the same node and the descendant
		classes will be close to zero as well.}
	\label{fig:wflow}
\end{figure}
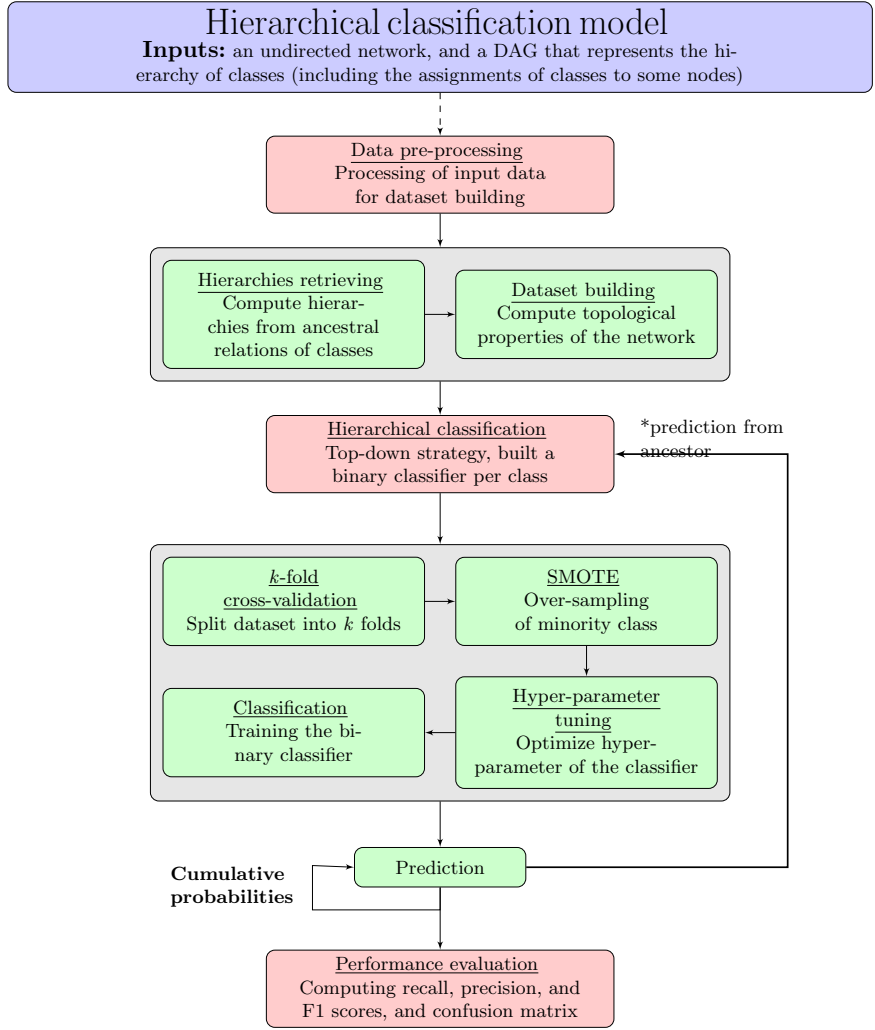

Given the network $\gG$ and the hierarchy tree $\gT$, the model is
built in a process comprising three stages. Figure~\ref{fig:wflow}
depicts the general approach.

\paragraph{\bf Stage 0: data pre-processing.} In this stage,
topological features of $\gG$ and $\gT$, and hierarchical information
in $\gT$ are readied and combined for supervised learning.

Classes that are too specific or too general are ignored in the
prediction to avoid overfitting and learning bias.
In the case study presented in Section~\ref{sec:gene}, a class is defined as
too specific or too general if it is associated to less than 5 or more than
300 genes, respectively~\cite{jiang-hbn-2008}. As a result, the input
hierarchy $\gT$ can be split into several sub-trees, each one representing a
sub-hierarchy $\gT'=(V_{\gT'},E_{\gT'})$ with $V_{\gT'}\subseteq V_{\gT}$ and
$E_{\gT'}\subseteq E_{\gT}$, over which the model is applied independently. 
That is, a sub-hierarchy $\gT'$ is a subset of the classes and ancestral
relations in $\gT$.
As a matter of fact, this situation arises in the case study presented in
Section~\ref{sec:gene}. Furthermore, each sub-hierarchy $\gT'$ is associated
to the subgraph of $\gG$ consisting of all nodes labeled with the root class
of $\gT'$, that is, each sub-hierarchy $\gT'$ is related to a different
subgraph $\gG'=(V_{\gG}', E_{\gG}')$ with $V_{\gG}'\subseteq V_{\gG}$ and
$E_{\gG}'\subseteq E_{\gG}$.
In this way,
sub-hierarchies are considered independent problems with smaller
inputs (\textit{\`a la} divide and conquer).

For each sub-hierarchy $\gT'$ in $\gT$, datasets are built based on two types
of topological properties, namely, hand-crafted features and node
embeddings. For the first type, properties of nodes $V_{\gG}'$ such as degree,
average neighbor degree, centrality, and eccentricity are
computed. 
Additionally, for each class $C$ in $\gT'$, two features are computed to
represent the probability of a node being associated to $C$ and its parent in
$\gH$ based on the information of the neighborhood. For $C$ and its parent, a
node and its neighbors, and the associations between the neighbors and both
classes, these new features represent the ratio between the number of
neighbors associated to each class and the total number of neighbors.
For the second type of properties, continuous representations
capturing the characteristics of the nodes in $\gG'$ (i.e. node
embeddings) are computed using node2vec~\cite{grover-node2vec-2016}.

\paragraph{\bf Stage 1: hierarchical classification.} This stage
comprises a top-down approach combining different supervised machine
learning techniques/tools. It builds prediction classifiers for each
sub-hierarchy $\gT'$ independently. The approach uses stratified
\textit{k}-fold cross-validation, the Synthetic Minority Over-sampling
Technique (SMOTE)~\cite{chawla-smote-2002}, hyper-parameter
tuning~\cite{bergstra-hpo-2012}, and a binary classifier,
(e.g. XGBoost~\cite{chen-xgboost-2016} or graph convolutional
networks~\cite{kipf-gcn-2017}).  These techniques are combined
sequentially in a pipeline, which is used iteratively from the root to
the leaves of each sub-hierarchy $\gT'$. Note that, since the top-down
approach builds a different classifier for each class in the
sub-hierarchy, the proposed model can be used for multi-class and
multi-label problems. As a result, nodes can be independently
associated to multiple classes.

The combination of the above-mentioned techniques/tools makes up the
core of the approach; and each technique has a different
objective. 
Stratified \textit{k}-fold aims to overcome overfitting by randomly
	selecting independent $k$ subsets of the dataset where the distribution of the
	labels is similar for all folds. In this approach, 5 folds are used for
	cross-validation, that is the train-test ratio is 80/20.
Over-sampling aims to overcome learning bias handling imbalanced datasets for
underrepresented classes. SMOTE synthesizes new examples of the minority class
from the existing ones.
Hyper-parameter tuning aims to improve the performance of the
prediction by optimizing parameters of the classifier such as, e.g.,
learning rate, number of estimators, and maximum depth of trees. 

Two types of classifiers were used; namely, the
XGBoost~\cite{chen-xgboost-2016} gradient boosting decision trees and graph
convolutional networks~\cite{kipf-gcn-2017}. XGBoost was chosen for
interpretability~\cite{elshawi-inter-2019,rudin-inter-2019} and graph
convolutional networks for state-of-the-art performance. In general, any other
binary classifier can be used in this stage.
The typical parameter values used for XGBoost classifiers are:
  \textit{gbtree} booster, area under Precision-Recall
  (\textit{aucpr}) evaluation metric, learning rate (\textit{eta}) of
  0.05, maximum tree depth (\textit{max\_depth}) of 6, subsample ratio
  (\textit{subsample}) of 0.9, and minimum sum of instance weight in a
  child (\textit{min\_child\_weight}) of 3.  For the graph
  convolutional networks, the implementation by~\cite{csiro-sg-2018}
  was used with the following parameters: 16 layers of 16 units each,
  RelU activation function, dropout rate of 50\%, learning rate of
  0.01, and binary cross-entropy loss function. Further details of the
  implementation can be founded in the repository
  \url{https://github.com/migueleci/node_classification}.

Classifiers for each class in $\gT'$ are built independently, so that there is
no relation between their predictions. Including information from the
ancestors of a class $C$ into its classifier is not enough to avoid
inconsistent predictions. For this reason, a correction mechanism is included
in this stage.
Since ensuring the true-path rule is key in the proposed approach, this stage
computes cumulative probabilities along the paths of classes in $\gT'$.
Namely, the probability of association between a node $v$ and $C$ is directly
related to the predicted probabilities of the node being associated to all
ancestors of $C$. Intuitively, the principle is as follows: if the probability
of association of $C$ to $v$ is close to zero, then the probability of
association for all descendant classes of $C$ to $v$ will be close to zero as
well. The main consequence of enforcing the principle is that the
classification computed from the cumulative probability satisfies the
true-path rule and removes the inconsistencies in the prediction.

\paragraph{\bf Stage 2: performance evaluation.} This stage comprises
the evaluation of the metrics used for measuring the prediction
performance of the classifiers. Performance evaluation focuses on
recall (true positive rate) and precision scores. It also evaluates
the precision-recall curve instead of the accuracy, loss, or ROC
curves. This is mainly because datasets are often imbalanced (w.r.t.
the positive class in a binary classification), thus both positive and
negative classes of the binary classifier need to be analyzed
separately. Recall and precision scores are computed from the
predicted cumulative probabilities as a function of the \emph{optimum
	threshold}, which is defined as the threshold that maximizes the F1
score from the precision-recall curve for the cumulative
probabilities.

\section{Gene Function Prediction}
\label{sec:gene}

This section presents a case study on the prediction of gene functions
(i.e., biological processes in which genes are involved) for the
\textit{Oryza sativa Japonica} rice variety. First, the problem of
predicting gene functions is introduced. Then, the results after
applying the approach proposed in Section~\ref{sec:model} to this
problem are described. The probabilistic approach, proposed
in~\cite{jiang-hbn-2008}, is used to compare the novel results.

\subsection{Gene Co-expression Networks}

High-throughput sequencing technologies have enabled the identification of
numerous genes and gene products. However, biological processes in which many
such genes are involved remain largely unknown (i.e., relations between genes
and biological processes have not been comprehensively validated through in
vivo experimentation)~\cite{ranganathan-bioinformatics-2019}.
Identifying the functions of genes is key to enhance the understanding
on how to characterize the genome of a particular organism. In
general, traditional in silico approaches to predict gene
functions consider each function as an independent class. The task is
generally defined as a binary classification problem based on gene
expression.

Genes (or gene products) can be associated to more than one biological process
and such processes may be related (e.g., by ancestral relationships). The
assignment of functions to genes obeys the true-path
rule~\cite{valentini-tpr-2009}. Consequently, efforts to predict whether a
gene is associated to a particular function should consider the ancestral
relations of that function. Ignoring such a hierarchical structure leads to
biological inconsistencies in the outcome of the prediction. On the contrary,
when a gene is associated to multiple biological processes without ancestral
relations, the prediction is done for each one of the functions independently.

\begin{figure}[tbph!]
	\centering
	\includegraphics[width=0.5\linewidth]{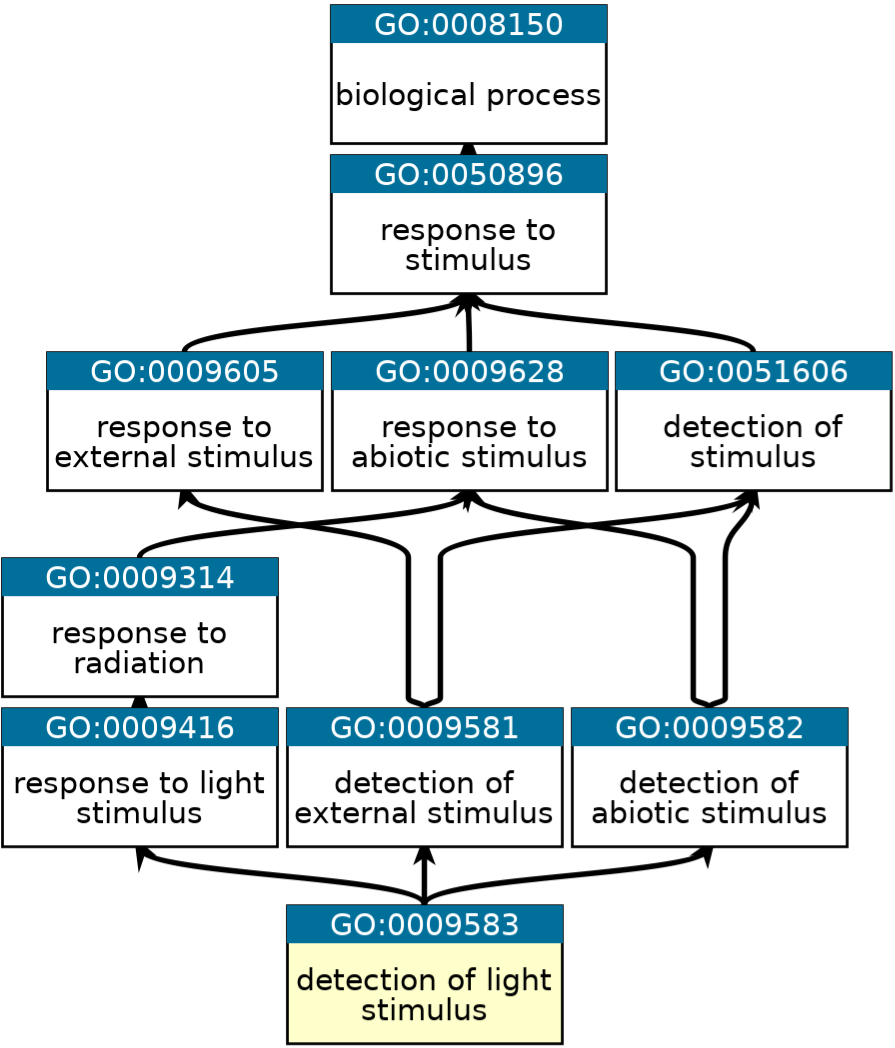}
	\caption{Hierarchy for the biological process \textit{detection of light
			stimulus}, represented as a DAG. Taken from QuickGO,
		\texttt{https://www.ebi.ac.uk/QuickGO}.}
	\label{fig:gotree}
\end{figure}

As an example, consider two biological processes in which a gene may
be involved: \textit{response to external stimulus} and
\textit{detection of light stimulus}. According to the hierarchy of
biological processes in Figure~\ref{fig:gotree}, the former function
is an ancestor of the latter~\cite{ashburner-go-2000}. By the
true-path rule, if a gene is associated to \textit{detection of light
	stimulus}, it must be associated to \textit{response to external
	stimulus}.

In addition, a common approach to integrate large volumes of
transcriptional data and synthesize the hierarchical structure of gene
functions is to characterize gene co-expression networks (GCNs). GCNs
have been used to infer biological processes and pathways based on
highly correlated expression patterns between
genes~\cite{oti-coexp-2008,vandam-disease-2017,vandepoele-arabidopsis-2009}.
It is well-known that co-expressed genes, i.e., genes with similar
expression profiles, tend to share the same function or be related to
the same regulatory
pathway~\cite{emamjomeh-gcnrec-2017,serin-gcn-2016,zhou-funcannot-2002}.

\subsection{Predicting Gene Functions in \textit{Oryza sativa Japonica}}

The goal of this case study is to predict gene functions, that is, the
biological processes in which some genes are involved. The problem is
tackled by using the model proposed in Section~\ref{sec:model} on the GNC of
\textit{Oryza sativa Japonica}~\cite{obayashi-atted2018-2018} and a
hierarchy of biological processes for this
organism~\cite{sakai-rapdb-2013}. The computational experiments
supporting the results in this section have been executed in a cluster
with 5 nodes, each one with 64GB of memory and a AMD
Opteron\texttrademark{} Processor 6376 with 64 CPU cores.

Formally, a gene co-expression network is represented as a undirected,
weighted graph $\gG=(V_G,E_G,f)$, built from empirical data, where
genes are represented by nodes $V_G$, edges $E_G$ denote co-expression
relationships, and the weight $f:E_G \to \mathbb{R}_{\geq 0}$ measures
the level of co-expression between genes. Additionally, the graph
$\gH=(V_H,E_H)$ is a directed acyclic graph (DAG) which represents the
hierarchical organization of biological processes, where $E_H$
represents the ancestral relations between functions. Genes are
associated to one or more biological processes through a function
$\phi : V_G \to 2^{V_H}$, where $V_H$ denotes the set of all
biological processes. The predictive model combines the existing set
of labels in $\phi$, topological properties of $\gG$ and the
hierarchical information of $\gH$ to obtain a new labeling function
$\phi'$ using the hierarchical multi-label classification approach. As
a result, the function $\phi'$ contains suggestions of previously
unidentified associations between genes and functions satisfying the
true-path rule.

The set of known associations between genes and functions used in this
work contains \numprint{19663} rice genes, \numprint{550813}
co-expression relations, \numprint{3743} biological processes,
\numprint{220598} assignments of functions to genes, and
\numprint{7185} ancestral relations of functions (all biological
processes belong to the same hierarchy). To avoid overfitting and
learning bias in the proposed model, only those functions associated
to more than 4 and at most 300 genes are
considered~\cite{jiang-hbn-2008}. Under this criterion,
\numprint{1938} functions (52\%) are used for prediction. As a result,
the function hierarchy breaks down into 27 sub-hierarchies, from which
12 correspond to isolated functions or small sub-hierarchies (fewer
than 7 functions). The 15 remaining sub-hierarchies are described in
Table~\ref{tab:hier}, sorted from the smallest to the largest in terms
of the number of functions $V_{\gT'}$ and number genes associated with
each of them.

\begin{table}[ht!]
	\centering
	\begin{tabular}{p{0.2cm} p{1.5cm} R{1cm} R{1.1cm} L{6.9cm}}
		\hline
		& Root & Func & Genes & Desc \\\hline\hline
		1 & GO:0040007 & 10 & 108 & growth \\
		2 & GO:0002376 & 11 & 131 & immune system process \\
		3 & GO:0051704 & 22 & 144 & multi-organism process \\
		4 & GO:0044419 & 37 & 777 & interspecies interaction between organisms \\
		5 & GO:0044085 & 50 & 377 & cellular component biogenesis \\
		6 & GO:0000003 & 72 & 648 & reproduction \\
		7 & GO:0006796 & 118 & \numprint{1270} & phosphate-containing compound metabolic process \\
		8 & GO:0032501 & 118 & \numprint{1043} & multicellular organismal process \\
		9 & GO:0032502 & 149 & \numprint{1063} & developmental process \\
		10 & GO:0016043 & 140 & 661 & cellular component organization \\
		11 & GO:0051179 & 164 & \numprint{1350} & localization \\
		12 & GO:0050896 & 261 & \numprint{3319} & response to stimulus \\
		13 & GO:0065007 & 485 & \numprint{2224} & biological regulation \\
		14 & GO:0008152 & 775 & \numprint{5862} & metabolic process \\
		15 & GO:0009987 & 925 & \numprint{5900} & cellular process \\
		\hline
	\end{tabular}
	\caption{Sub-hierarchies generated for the gene co-expression network of
		\textit{Oryza sativa Japonica}}
	\label{tab:hier}
\end{table}

The prediction performance of the proposed approach is compared with
the HBN model presented in~\cite{jiang-hbn-2008}. The HBN model uses a
\textit{top-down} approach that integrates relational data of
protein-protein interaction network (PPI) with the hierarchical data
of biological processes with the objective of predicting protein
functions. For this case study, the HBN model is adapted to the
problem of predicting gene functions based on GCNs. To predict the
probability of a gene $g$ being associated to function $A$, the local
neighborhood information of $g$ in the GCN and the ancestors
of $A$ in the hierarchy are considered. The HBN model computes the
probability of gene $g$ being associated to function $A$ obeying the
true-path rule.

\begin{figure}[tbph!]
	\centering
	\includegraphics[width=0.95\linewidth]{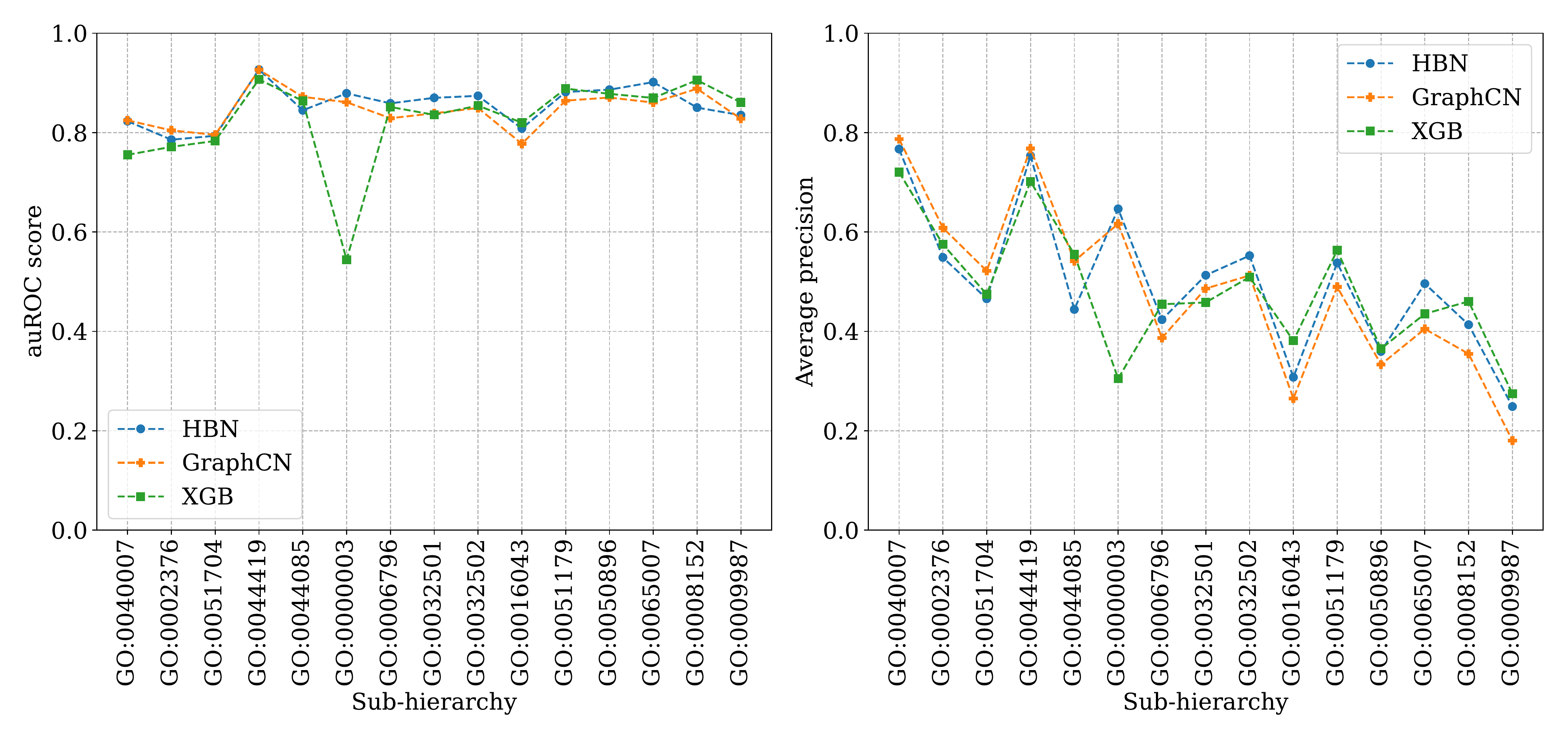}
	\caption{Prediction performance of the hierarchical multi-label
		classification approach with XGBoost (XGB) and graph convolutional
		network (GraphCN) classifiers, and the probabilistic model (HBN) for
		the 15 sub-hierarchies of \textit{Oriza sativa Japonica}.
		Performance is measured with area under the ROC curve and the
		average precision score. Note that the notation used for the graph
		convolutional network is GraphCN to distinguish it from the gene
		co-expression network (GCN).}
	\label{fig:score}
\end{figure}

The figures in this section show the mean performance for the
  proposed approach and the HBN model between multiple experiments, in
  which each experiment represents the mean performance between the
  $k$ folds used for cross-validation. In all of them, the variation
  (error bar o standard deviation) is not included because it is
  negligible (and can add visual noise to the plots).
Figure~\ref{fig:score} illustrates the performance of the proposed
approach using XGBoost (XGB) and graph convolutional network (GraphCN)
classifiers and the HBN model measured with the area under the ROC
curve and the average precision score. Note that their performance
seem to be similar in most sub-hierarchies and it is not possible to
conclude which one performs better from Figure~\ref{fig:score}.
However, since only biological processes associated to more than 4 and
less than 300 genes are considered (less than 2\% of the genes in the
GCN), datasets generated for the filtered biological processes are
highly imbalanced. For this reason, the area under the ROC curve is
not suitable for the case study (this measured is biased for the
over-represented class in the classification task), and the analysis
should focus on other metrics such as recall and F1 score instead.

\begin{figure}[tbph!]
	\centering
	\includegraphics[width=0.95\linewidth]{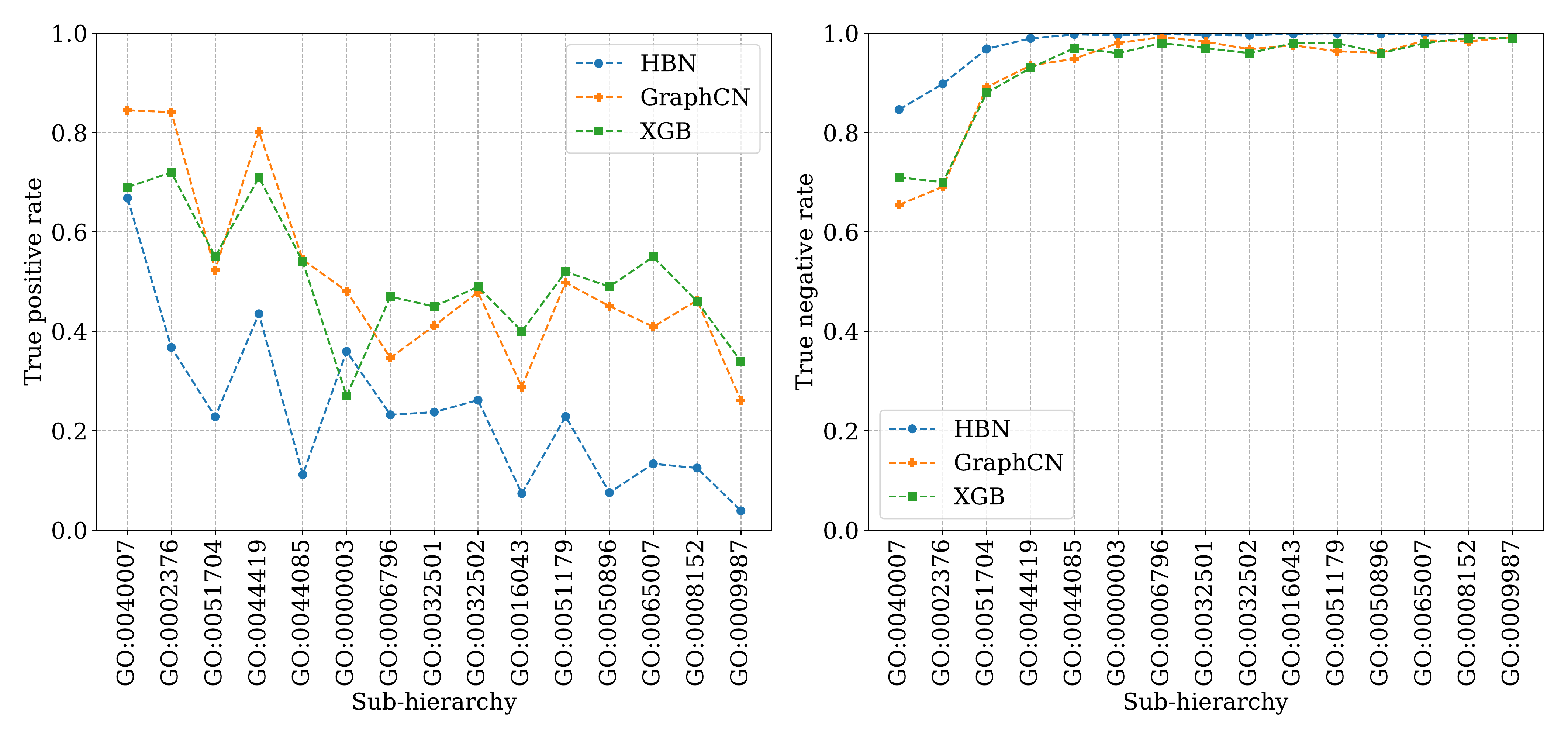}
	\caption{True positive rate (or recall) and true negative rate of the
		hierarchical multi-label classification approach with XGBoost (XGB)
		and graph convolutional network (GraphCN) classifiers, and the HBN
		model for the 15 sub-hierarchies generated for \textit{Oryza sativa
			Japonica}.}
	\label{fig:tpr-tnr}
\end{figure}

An outstanding difference between the performance of the proposed
approach and the HBN model is observed when the confusion matrices are
analyzed.  Figure~\ref{fig:tpr-tnr} shows the true positive rate (or
the measure of recall) and the true negative rate for the 15
sub-hierarchies. Note that the true positive rate of the proposed
approach is higher than the HBN model for most of the sub-hierarchies,
whereas the true negative rate of the HBN model is higher for all
  sub-hierarchies. However, the HBN model is biased for the negative
  class because the probability predicted by the HBN model for most of
  the associations between genes and functions is close to zero.
As the datasets are highly imbalanced, the performance in terms of the
positive class are key to determine which approach is adequate. Recall that a
dataset is said to be \emph{imbalanced} for binary classification if one of
the classes is under-represented in relation to the other one, i.e., the
number of instances related to one class is much higher than the number of
instances related to the other. For example, a dataset with \numprint{1000}
instances that has 900 negative and 100 positive samples is imbalanced.

\begin{figure}[tbph!]
	\centering
	\includegraphics[width=0.7\linewidth]{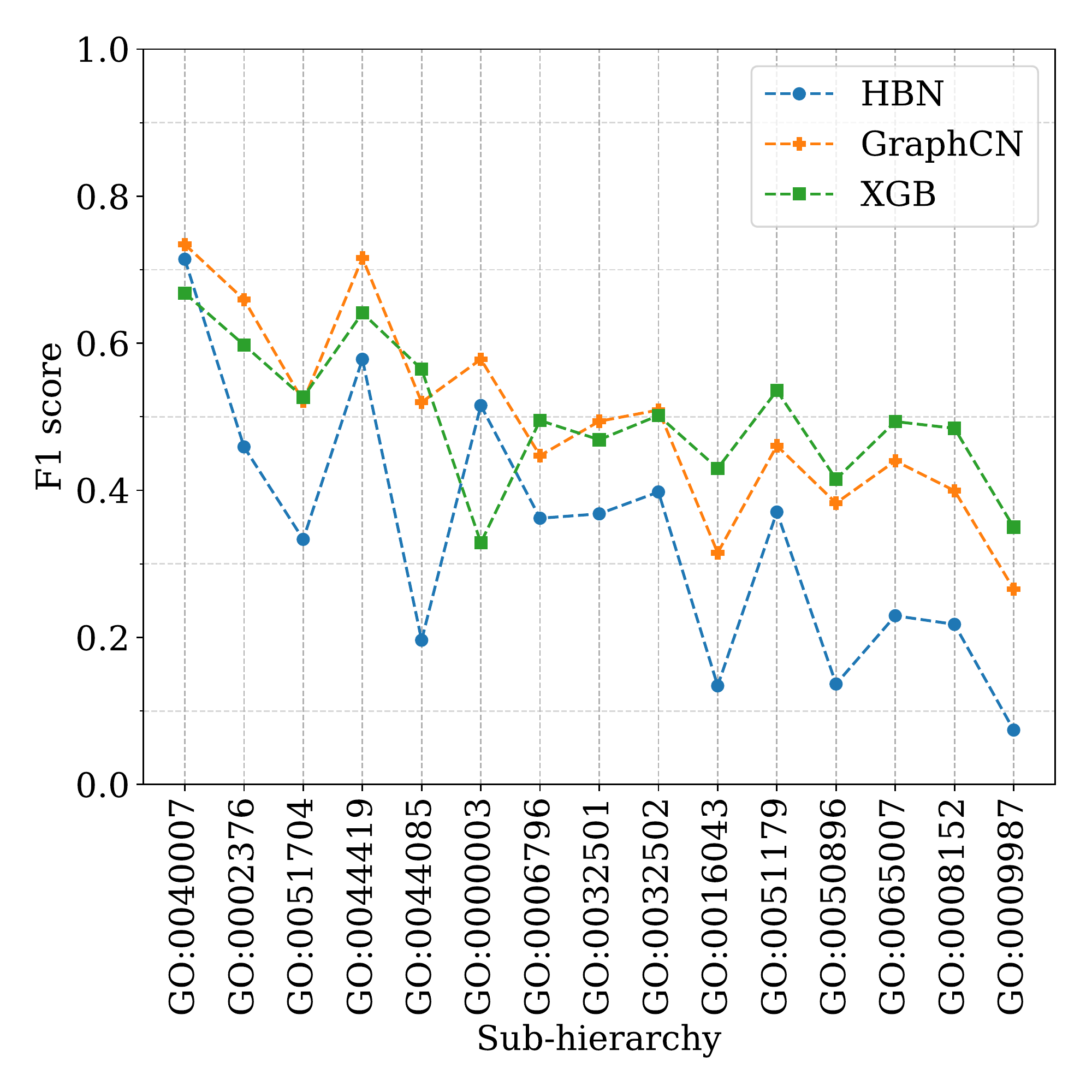}
	\caption{F1 score of the hierarchical multi-label classification
		approach with XGBoost (XGB) and graph convolutional network (GraphCN)
		classifiers, and the HBN model for the 15 sub-hierarchies generated
		for \textit{Oryza sativa Japonica}.}
	\label{fig:f1s}
\end{figure}

The true positive rate illustrated in Figure~\ref{fig:tpr-tnr} shows
that the proposed approach outperforms the HBN model in the
identification of the (positive) associations between genes and
functions.
The performance varies between XGBoost and graph convolutional networks, but
both classifiers have better overall performance than the HBN model. The
results suggest that graph convolutional networks are better for small
sub-hierarchies, while XGBoost is better for larger ones.
Even though the true negative rate of the HBN model is close to 1 for
all sub-hierarchies, as illustrated on Figure~\ref{fig:tpr-tnr}, the
performance of the proposed approach in terms of the average of both
recall and precision (i.e., F1 score) is better than the HBN
model. Figure~\ref{fig:f1s} presents the F1 score of the proposed
approach and the HBN model for the 15 sub-hierarchies.
In this case study there is no observable correlation between the
  size/depth/span of a hierarchy and the prediction performance,
  according to the experiments. This is coherent with the overall
  computational complexity of the algorithms. On the other hand, there
  is no experimental evidence to suggest that some degree of
  correlation exists between the number of label nodes and the
  prediction performance. However, these observations need to be
  further investigated with other case studies.

\begin{figure}[tbph!]
	\centering
	\includegraphics[width=0.7\linewidth]{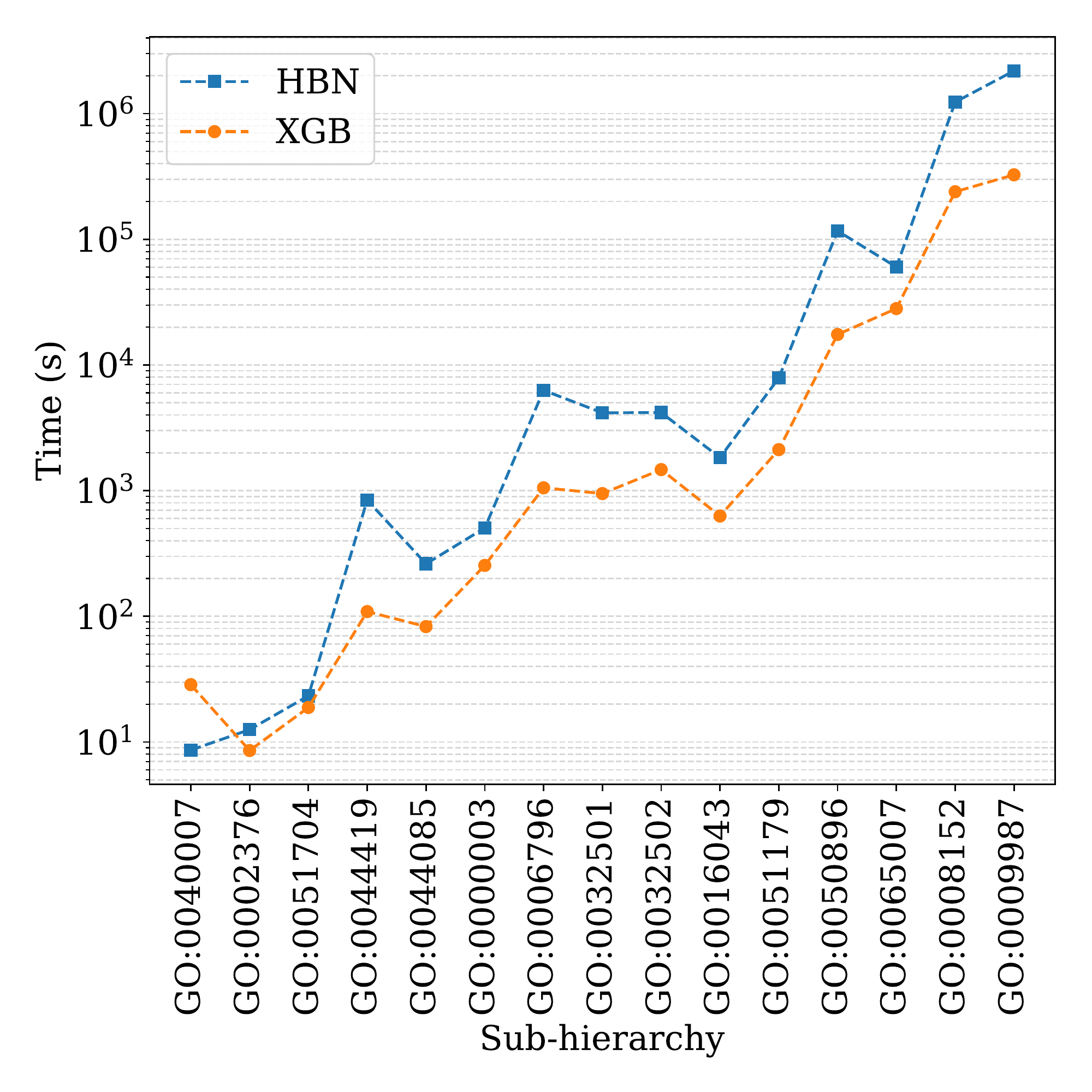}
	\caption{Execution time of the hierarchical multi-label
		classification approach with XGBoost (XGB) classifier and the HBN
		model for the prediction of the 15 sub-hierarchies. The execution
		time is measured in seconds and plotted in logarithmic scale.}
	\label{fig:time}
\end{figure}

Finally, the execution time of the proposed approach and the HBN model
is illustrated on Figure~\ref{fig:time}. 
The execution time for the graph convolutional network classifier is not
included because the experiments were executed on CPUs rather than GPU. It is
known that neural networks run much faster on GPUs; thus, it would not be fair
to make a comparison with the available data.
Note that the execution time is measured in seconds and plotted on a
logarithmic scale. Except for the smallest sub-hierarchy (GO:0040007),
the execution time of the proposed approach, using XGBoost classifier,
is better than the HBN model. On average, the execution time of the
HBN model is approximately 4 times as much of the proposed approach.
	
\section{Conclusion and Future Work}
\label{sec:concl}

By combining different techniques from machine learning, the
hierarchical multi-label classification model presented in this paper
introduces an approach to address the node classification problem for
scenarios in which nodes can have attributes obeying a hierarchical
organization. Taken into account hierarchical dependencies is shown to
be a key aspect for obtaining more consistent predictions that satisfy
the true-path rule.

A baseline comparison between the proposed approach using two
different classification methods, namely, gradient boosting decision
trees and graph convolutional networks, and the HBN model introduced
by~\cite{jiang-hbn-2008} is presented. Both approaches are applied to
the problem of predicting gene function on the variety of rice
\textit{Oryza sativa Japonica}.  The proposed hierarchical multi-label
classification approach outperforms the HBN model in two
aspects. First, using topological information of the network is a key
feature to obtain the overall best performance of the prediction. In
such setting, the true positive rates of the proposed approach are
significantly higher than the HBN model, whereas the true negative
rates yield similar values (close to one). This result suggests that
the proposed approach can lead to good prediction of associations
between genes and functions in \textit{Oryza sativa Japonica} and,
potentially, in other organisms.

For scenarios in which the classes of the hierarchy are
under-represented, i.e., datasets are imbalanced, it is important to
center the performance analysis on metrics that are not biased by the
imbalanced dataset. Such metrics include the true positive rate (or
the measure of recall), the true negative rate, and the
F1-score. Other widely-used metrics, like the area under ROC curve and
the measure of average precision, are misleading for evaluating the
performance of a classifier under such conditions.

Second, the execution time of the proposed approach for the XGBoost
classifier is, on average, 4 times better than that of the HBN model.
The reduction in computational cost of the proposed top-down approach
can be attributed to the fact that it predicts the probability of
associations between a class and every node of the network at the same
time. Also, the efficient computation of the DAG into a tree helps in
making the proposed approach relevant to analyze larger networks and
hierarchies.

Finally, although the performance of the proposed approach is
promising, it requires to gather sufficient information from node
classes, which in some cases is incomplete or unavailable. For
example, information about gene functions is limited for many genes
and gene products. For some organisms there is no such information
available at all. The shortage of information may lead to over-fitting
or learning bias in the approach, and consequently to misleading
conclusions. Including other networks as additional sources of
information for the classification problem seems to be interesting for
future work. Other networks can be added with the help of transfer
learning techniques. For example, by creating new features that
aggregate the information extracted from other networks that can be
integrated in the proposed approach as additional input to improve the
prediction performance. Furthermore, other approaches such as
semi-supervised and transductive learning can also be considered for
future work to handle the amount of data required for training.

\section*{Abbreviations}
\begin{description}
	\item[DAG:] Directed Acyclic Graph
	\item[GO:] Gene Ontology 
	\item[GCN:] Gene Co-expression Network
	\item[HBN:] Hierarchical Binomial-Neighborhood
	\item[HMC:] Hierarchical Multi-label Classification
	\item[ROC:] Receiver Operating Characteristic
	\item[SMOTE:] Synthetic Minority Over-sampling Technique
\end{description}

\section*{Availability of data and materials}
The datasets analyzed for the current study are publicly available from
different sources. They can be found in the following locations:

\begin{itemize}	
	\item Gene co-expression data of \textit{Oryza sativa Japonica} is available on ATTED-II~\cite{obayashi-atted2018-2018}.
	\item Functional data of rice genes is available on~\cite{sakai-rapdb-2013} and~\cite{kurata-oryzabase-2006}.
\end{itemize}

The data collected, cleaned, and processed from the above sources as
used in the case study can be requested to the authors.

A workflow implementation is publicly available:
\begin{itemize}
	\item Project name: Node Classification
	\item Project home page: \url{https://github.com/migueleci/node_classification}
	\item Operating system(s): platform independent.
	\item Programming language: Python 3.
	\item Other requirements: None.
	\item License: GNU GPL v3.
\end{itemize}

\section*{Funding}
This work was funded by the OMICAS program: Optimización Multiescala
In-silico de Cultivos Agrícolas Sostenibles (Infraestructura y
Validación en Arroz y Caña de Azúcar), anchored at the Pontificia
Universidad Javeriana in Cali and funded within the Colombian
Scientific Ecosystem by The World Bank, the Colombian Ministry of
Science, Technology and Innovation, the Colombian Ministry of
Education and the Colombian Ministry of Industry and Turism, and
ICETEX, under GRANT ID: FP44842-217-2018.

\bibliographystyle{abbrv}
\bibliography{main}

\begin{thebibliography}{10}

\bibitem{abu-ngcn-2019}
S.~Abu-El-Haija, B.~Perozzi, A.~Kapoor, and J.~Lee.
\newblock N-gcn: Multi-scale graph convolutionfor semi-supervised node
  classification.
\newblock In {\em Conference on Uncertainty in Artificial Intelligence (UAI)},
  2019.

\bibitem{ashburner-go-2000}
M.~Ashburner, C.~A. Ball, J.~A. Blake, D.~Botstein, H.~Butler, J.~M. Cherry,
  A.~P. Davis, K.~Dolinski, S.~S. Dwight, J.~T. Eppig, M.~A. Harris, D.~P.
  Hill, L.~{Issel-Tarver}, A.~Kasarskis, S.~Lewis, J.~C. Matese, J.~E.
  Richardson, M.~Ringwald, G.~M. Rubin, and G.~Sherlock.
\newblock Gene {{Ontology}}: Tool for the unification of biology.
\newblock {\em Nature Genetics}, 25(1):25--29, May 2000.

\bibitem{bergstra-hpo-2012}
J.~Bergstra and Y.~Bengio.
\newblock Random {{Search}} for {{Hyper}}-{{Parameter Optimization}}.
\newblock {\em Journal of Machine Learning Research}, 13(10):281--305, 2012.

\bibitem{bhagat-nodecl-2011}
S.~Bhagat, G.~Cormode, and S.~Muthukrishnan.
\newblock Node {{Classification}} in {{Social Networks}}.
\newblock In C.~C. Aggarwal, editor, {\em Social Network Data Analytics}, pages
  115--148. {Springer US}, {Boston, MA}, 2011.

\bibitem{bi-hmlc-2011}
W.~Bi and J.~T. Kwok.
\newblock Multi-label classification on tree- and dag-structured hierarchies.
\newblock In {\em Proceedings of the 28th International Conference on
  International Conference on Machine Learning}, ICML'11, page 17–24,
  Madison, WI, USA, 2011. Omnipress.

\bibitem{chawla-smote-2002}
N.~V. Chawla, K.~W. Bowyer, L.~O. Hall, and W.~P. Kegelmeyer.
\newblock {{SMOTE}}: {{Synthetic Minority Over}}-sampling {{Technique}}.
\newblock {\em Journal of Artificial Intelligence Research}, 16:321--357, June
  2002.

\bibitem{chen-gfp-2021}
Q.~Chen, Y.~Li, K.~Tan, Y.~Qiao, S.~Pan, T.~Jiang, and Y.-P.~P. Chen.
\newblock Network-based methods for gene function prediction.
\newblock {\em Briefings in Functional Genomics}, 20(4):249--257, July 2021.

\bibitem{chen-xgboost-2016}
T.~Chen and C.~Guestrin.
\newblock {{XGBoost}}: {{A Scalable Tree Boosting System}}.
\newblock {\em Proceedings of the 22nd ACM SIGKDD International Conference on
  Knowledge Discovery and Data Mining}, pages 785--794, 2016.

\bibitem{csiro-sg-2018}
C.~Data61.
\newblock Stellargraph machine learning library.
\newblock \url{https://github.com/stellargraph/stellargraph}, 2018.

\bibitem{dimitrovski-hmlc-2010}
I.~Dimitrovski, D.~Kocev, S.~Loskovska, and S.~D{\v z}eroski.
\newblock Detection of {{Visual Concepts}} and {{Annotation}} of {{Images Using
  Ensembles}} of {{Trees}} for {{Hierarchical Multi}}-{{Label Classification}}.
\newblock In D.~{\"U}nay, Z.~{\c C}ataltepe, and S.~Aksoy, editors, {\em
  Recognizing {{Patterns}} in {{Signals}}, {{Speech}}, {{Images}} and
  {{Videos}}}, volume 6388, pages 152--161. {Springer Berlin Heidelberg},
  {Berlin, Heidelberg}, 2010.

\bibitem{elshawi-inter-2019}
R.~Elshawi, M.~H. {Al-Mallah}, and S.~Sakr.
\newblock On the interpretability of machine learning-based model for
  predicting hypertension.
\newblock {\em BMC Medical Informatics and Decision Making}, 19(1):146, July
  2019.

\bibitem{emamjomeh-gcnrec-2017}
A.~Emamjomeh, E.~Saboori~Robat, J.~Zahiri, M.~Solouki, and P.~Khosravi.
\newblock Gene co-expression network reconstruction: A review on computational
  methods for inferring functional information from plant-based expression
  data.
\newblock {\em Plant Biotechnology Reports}, 11(2):71--86, Apr. 2017.

\bibitem{go-go-2019}
{Gene Ontology Consortium}.
\newblock The {{Gene Ontology Resource}}: 20 years and still {{GOing}} strong.
\newblock {\em Nucleic Acids Research}, 47(D1):D330--D338, Jan. 2019.

\bibitem{grover-node2vec-2016}
A.~Grover and J.~Leskovec.
\newblock node2vec: Scalable feature learning for networks, 2016.

\bibitem{hamilton-induct-2017}
W.~L. Hamilton, R.~Ying, and J.~Leskovec.
\newblock Inductive representation learning on large graphs.
\newblock In {\em Proceedings of the 31st International Conference on Neural
  Information Processing Systems}, {{NIPS}}'17, pages 1025--1035, {Red Hook,
  NY, USA}, Dec. 2017. {Curran Associates Inc.}

\bibitem{jiang-hbn-2008}
X.~Jiang, N.~Nariai, M.~Steffen, S.~Kasif, and E.~D. Kolaczyk.
\newblock Integration of relational and hierarchical network information for
  protein function prediction.
\newblock {\em BMC Bioinformatics}, 9(1):350, 2008.

\bibitem{khan-binary-2010}
S.~S. Khan and M.~G. Madden.
\newblock A {{Survey}} of {{Recent Trends}} in {{One Class Classification}}.
\newblock In L.~Coyle and J.~Freyne, editors, {\em Artificial {{Intelligence}}
  and Cognitive Science}, volume 6206, pages 188--197. {Springer Berlin
  Heidelberg}, {Berlin, Heidelberg}, 2010.

\bibitem{kipf-gcn-2017}
T.~N. Kipf and M.~Welling.
\newblock Semi-supervised classification with graph convolutional networks.
\newblock In {\em International Conference on Learning Representations (ICLR)},
  2017.

\bibitem{knuth-programming-1997}
D.~E. Knuth.
\newblock {\em The Art of Computer Programming}.
\newblock {Addison-Wesley}, {Reading, Mass}, 3rd ed edition, 1997.

\bibitem{kurata-oryzabase-2006}
N.~Kurata and Y.~Yamazaki.
\newblock Oryzabase. {{An Integrated Biological}} and {{Genome Information
  Database}} for {{Rice}}.
\newblock {\em Plant Physiology}, 140(1):12--17, Jan. 2006.

\bibitem{makrodimitris-afp-2020}
S.~Makrodimitris, R.~C. H.~J. {van Ham}, and M.~J.~T. Reinders.
\newblock Automatic {{Gene Function Prediction}} in the 2020's.
\newblock {\em Genes}, 11(11):1264, Oct. 2020.

\bibitem{mills-multiclass-2021}
P.~Mills.
\newblock Solving for multi-class: A survey and synthesis.
\newblock {\em arXiv:1809.05929 [cs, stat]}, Jan. 2021.

\bibitem{obayashi-atted2018-2018}
T.~Obayashi, Y.~Aoki, S.~Tadaka, Y.~Kagaya, and K.~Kinoshita.
\newblock {{ATTED}}-{{II}} in 2018: {{A Plant Coexpression Database Based}} on
  {{Investigation}} of the {{Statistical Property}} of the {{Mutual Rank
  Index}}.
\newblock {\em Plant and Cell Physiology}, 59(1):e3--e3, Jan. 2018.

\bibitem{oti-coexp-2008}
M.~Oti, J.~{van Reeuwijk}, M.~A. Huynen, and H.~G. Brunner.
\newblock Conserved co-expression for candidate disease gene prioritization.
\newblock {\em BMC Bioinformatics}, 9(1):208, 2008.

\bibitem{prajapati-multilabel-2012}
P.~Prajapati, A.~Thakkar, and A.~Ganatra.
\newblock A survey and current research challenges in multi-label
  classification methods.
\newblock {\em International Journal of Soft Computing and Engineering
  (IJSCE)}, 2(1):248--252, 2012.

\bibitem{ramirez-corona-hmlc-2016}
M.~{Ram{\'i}rez-Corona}, L.~E. Sucar, and E.~F. Morales.
\newblock Hierarchical multilabel classification based on path evaluation.
\newblock {\em International Journal of Approximate Reasoning}, 68:179--193,
  Jan. 2016.

\bibitem{ranganathan-bioinformatics-2019}
S.~Ranganathan, M.~R. Gribskov, K.~Nakai, and C.~Sch{\"o}nbach.
\newblock {\em Encyclopedia of Bioinformatics and Computational Biology}.
\newblock {Elsevier}, 2019.
\newblock OCLC: 1052465484.

\bibitem{rudin-inter-2019}
C.~Rudin.
\newblock Stop explaining black box machine learning models for high stakes
  decisions and use interpretable models instead.
\newblock {\em Nature Machine Intelligence}, 1(5):206--215, May 2019.

\bibitem{sakai-rapdb-2013}
H.~Sakai, S.~S. Lee, T.~Tanaka, H.~Numa, J.~Kim, Y.~Kawahara, H.~Wakimoto,
  C.-c. Yang, M.~Iwamoto, T.~Abe, Y.~Yamada, A.~Muto, H.~Inokuchi, T.~Ikemura,
  T.~Matsumoto, T.~Sasaki, and T.~Itoh.
\newblock Rice {{Annotation Project Database}} ({{RAP}}-{{DB}}): {{An
  Integrative}} and {{Interactive Database}} for {{Rice Genomics}}.
\newblock {\em Plant and Cell Physiology}, 54(2):e6--e6, Feb. 2013.

\bibitem{serin-gcn-2016}
E.~A.~R. Serin, H.~Nijveen, H.~W.~M. Hilhorst, and W.~Ligterink.
\newblock Learning from {{Co}}-expression {{Networks}}: {{Possibilities}} and
  {{Challenges}}.
\newblock {\em Frontiers in Plant Science}, 7, Apr. 2016.

\bibitem{silla-hierarchy-2011}
C.~N. Silla and A.~A. Freitas.
\newblock A survey of hierarchical classification across different application
  domains.
\newblock {\em Data Mining and Knowledge Discovery}, 22(1-2):31--72, Jan. 2011.

\bibitem{valentini-tpr-2009}
G.~Valentini.
\newblock True {{Path Rule Hierarchical Ensembles}}.
\newblock In J.~A. Benediktsson, J.~Kittler, and F.~Roli, editors, {\em
  Multiple Classifier Systems}, volume 5519, pages 232--241. {Springer Berlin
  Heidelberg}, {Berlin, Heidelberg}, 2009.

\bibitem{vandam-disease-2017}
S.~{van Dam}, U.~V{\~o}sa, A.~{van der Graaf}, L.~Franke, and J.~P. {de
  Magalh{\~a}es}.
\newblock Gene co-expression analysis for functional classification and
  gene\textendash{}disease predictions.
\newblock {\em Briefings in Bioinformatics}, page bbw139, Jan. 2017.

\bibitem{vandepoele-arabidopsis-2009}
K.~Vandepoele, M.~Quimbaya, T.~Casneuf, L.~De~Veylder, and Y.~{Van de Peer}.
\newblock Unraveling {{Transcriptional Control}} in {{Arabidopsis Using}}
  cis-{{Regulatory Elements}} and {{Coexpression Networks}}.
\newblock {\em Plant Physiology}, 150(2):535--546, June 2009.

\bibitem{xiao-gnn-2021}
S.~Xiao, S.~Wang, Y.~Dai, and W.~Guo.
\newblock Graph neural networks in node classification: Survey and evaluation.
\newblock {\em Machine Vision and Applications}, 33(1):4, Nov. 2021.

\bibitem{zhou-funcannot-2002}
X.~Zhou, M.-C.~J. Kao, and W.~H. Wong.
\newblock Transitive functional annotation by shortest-path analysis of gene
  expression data.
\newblock {\em Proceedings of the National Academy of Sciences},
  99(20):12783--12788, Oct. 2002.

\end{thebibliography}

\end{document}